\PassOptionsToPackage{table,dvipsnames}{xcolor} % Pass options FIRST
\documentclass[letterpaper, 10pt, conference]{ieeeconf} % Document class AFTER
\usepackage{hyperref}
\usepackage{graphicx}
\usepackage{url}
\usepackage{multirow}
\usepackage{amsmath}
\usepackage{amssymb,amsfonts}
\usepackage{booktabs}
\usepackage{framed,multirow}
\usepackage{makecell}
\usepackage{arydshln}
\usepackage{soul}
\usepackage{adjustbox}
\usepackage{orcidlink}
\usepackage{algorithm}
\usepackage{algpseudocode}
\usepackage{float} 
\usepackage{xcolor} 

\title{\LARGE \bf
Action Recognition in Real-World Ambient Assisted Living Environment
}

\author{Vincent Gbouna Zakka$^{1*}$ Zhuangzhuang Dai $^{2}$ Luis J. Manso $^{3}$\\
\normalsize $^{1,2,3}$School of Computer Science and Digital Technologies, Aston University, Birmingham, B4 7ET, United Kingdom\\
{$^{*}$Corresponding address: \tt\small vzakk22@aston.ac.uk}
}

\begin{document}

\maketitle

\begin{abstract}
The growing ageing population and their preference to maintain independence by living in their own homes require proactive strategies to ensure safety and support. Ambient Assisted Living (AAL) technologies have emerged to facilitate ageing in place by offering continuous monitoring and assistance within the home. Within AAL technologies, action recognition plays a crucial role in interpreting human activities and detecting incidents like falls, mobility decline, or unusual behaviours that may signal worsening health conditions. However, action recognition in practical AAL applications presents challenges, including occlusions, noisy data, and the need for real-time performance. While advancements have been made in accuracy, robustness to noise, and computation efficiency, achieving a balance among them all remains a challenge. To address this challenge, this paper introduces the Robust and Efficient Temporal Convolution network (RE-TCN), which comprises three main elements: Adaptive Temporal Weighting (ATW), Depthwise Separable Convolutions (DSC), and data augmentation techniques. These elements aim to enhance the model's accuracy, robustness against noise and occlusion, and computational efficiency within real-world AAL contexts. RE-TCN outperforms existing models in terms of accuracy, noise and occlusion robustness, and has been validated on four benchmark datasets: NTU RGB+D 60, Northwestern-UCLA, SHREC'17, and DHG-14/28. The code is publicly available at: \href{https://github.com/Gbouna/RE-TCN}{https://github.com/Gbouna/RE-TCN}
\end{abstract}

\section{Introduction}
\label{s:introduction}
\noindent
According to the United Nations, the number of individuals aged 65 years or older is projected to double by 2050, reaching approximately 1.5 billion worldwide~\cite{unitednations2019}. This demographic shift presents significant challenges for healthcare systems, economics, and society at large~\cite{Jones}.

Despite these challenges, most older adults prefer to \textit{age in place}, desiring to live independently in their own homes rather than relocating to assisted living facilities or nursing homes~\cite{Bosch}. However, enabling ageing in place requires proactive measures to ensure safety and support~\cite{Lette}, especially as older adults become more susceptible to health risks such as falls --which are the leading cause of injury-related deaths. In this context, Ambient Assisted Living (AAL) technologies support ageing in place by providing continuous monitoring and assistance within the home environment~\cite{Stephanie,Bennasar}. Among the various components of AAL systems, action recognition plays a crucial role, enabling the system to detect and classify human actions, including events such as falls, mobility decline, or abnormal behaviours that may indicate health deterioration~\cite{Ranieri}.

Different sensing technologies are common in AAL systems. Wearable devices, such as accelerometers and gyroscopes, are used to monitor movements and detect falls~\cite{Vincenzo,Bennasar,Khimraj}. While wearables provide continuous monitoring, they rely heavily on user compliance causing discomfort, which can be problematic for older individuals, especially those with cognitive impairments~\cite{Vincenzo}. Audio-based systems can monitor sound patterns and movement within the home~\cite{Cristina,Zull}. However, the accuracy of these systems is degraded in the presence of background noise~\cite{Zull}. Recently, computer vision-based approaches have gained popularity for action recognition in AAL due to their ability to capture rich visual data of human movements and interactions~\cite{Vishwakarma}. 

While many vision-based approaches rely on traditional RGB video data for action recognition, this raises privacy concerns due to the intrusive nature of capturing full-colour video of individuals in their homes~\cite{Vishwakarma}. Moreover, RGB-based systems are computationally expensive, requiring significant processing power to handle large volumes of data generated by continuous video streams~\cite{Cailing}. 

A promising alternative is skeleton-based action recognition, which uses skeleton data extracted from images to represent human motion rather than using raw image data~\cite{Liliana,Fanuel}. This approach offers several advantages in the context of AAL systems. First, skeleton-based data preserves visual privacy by abstracting the human figure into a set of joints, eliminating the need to capture identifiable features such as facial details or body appearance~\cite{Akerkar,Htoo}. Furthermore, skeleton data is compact and lightweight, making it computationally more efficient to process than RGB video. This is an advantage for real-time action recognition tasks in resource-constrained environments~\cite{Gbouna,Wen,Yujian,Bingyi}. 

\begin{figure*}[!tb]
  \centering  \includegraphics[width=1\linewidth]{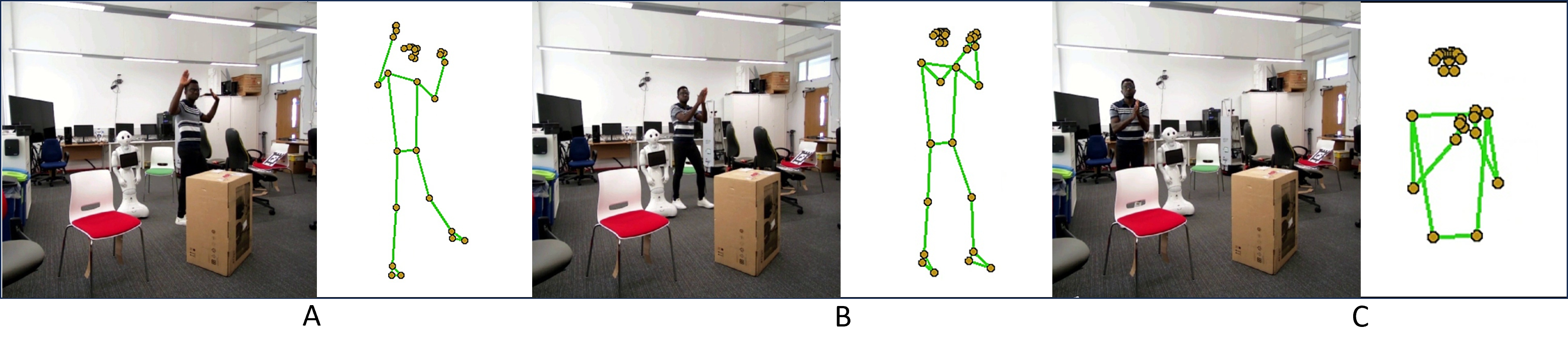}
  \caption{Challenges in a real-world environment: A) represents data with noise and occlusion, B) represents relatively clean data, and C) represents data with occlusion}
  \label{fig:opening_figure}
\end{figure*}

However, skeleton-based action recognition faces several challenges in real-world AAL settings as shown in Fig~\ref{fig:opening_figure}. One of the primary issues is dealing with occlusions, where body parts may be hidden from view due to obstacles or suboptimal camera placement~\cite{Zhenjie}. Another challenge is noisy data resulting from camera inaccuracies or motion artefacts~\cite{Tabkhi}. To address these challenges, various methods have been proposed, focusing on enhancing the robustness~\cite{yoon2022predictively,Tabkhi,Zhenjie}, accuracy~\cite{cheng2020skeleton}, and efficiency~\cite{song2020stronger,lin2019tsm} of skeleton-based systems. While progress has been made in specific areas, such as robustness, accuracy, or real-time performance, a solution that balances all three remains a challenge. 

This paper proposes Robust and Efficient Temporal Convolution Network (RE-TCN) that builds on a state-of-the-art model Temporal Decoupling Graph Convolution Neural Network (TD-GCN)~\cite{Xinshun} by incorporating three key components: Adaptive Temporal Weighting (ATW), Depthwise Separable Convolutions (DSC) and data augmentation techniques. These components are designed to improve both the accuracy, robustness to noise and occlusion, and computational efficiency of the proposed model in a real-world AAL environment. ATW enhances the model's ability to focus on the most informative frames in an action sequence. It dynamically assigns different levels of importance to each frame, allowing the model to prioritise key moments in the sequence. DSC decomposes the convolution into depthwise and pointwise convolutions, significantly reducing the number of parameters and operations required to process the input skeleton data. The data augmentation techniques were designed to enhance the model's robustness in real-world environments.

We conducted extensive experiments on four benchmark datasets: NTU RGB+D 60~\cite{7780484}, Northwestern-UCLA~\cite{wang2014crossview}, SHREC17~\cite{LI20151}, and DHG-14/28~\cite{de2016skeleton} to evaluate the effectiveness of the proposal. 
RE-TCN outperforms existing methods in terms of computational efficiency, accuracy, and robustness to noise and occlusion.

\section{Related Works}\label{s:Related_works}
\noindent\subsection{Action Recognition in Ambient Assisted Living}
There are two common approaches to Human Activity Recognition (HAR), rule-based and data-driven. 
Rule-based approaches employ thresholds for triggering alerts of potentially harmful or dangerous events~\cite{Ricardo}.
The second approach leverages machine learning algorithms for HAR~\cite{Rashidi}. These methods can learn complex patterns from data, enabling more accurate recognition of activities. These methods are implemented using various sensors, generally categorised into wearable and non-wearable sensors. 

Inertial sensors are the most common wearable sensors for HAR in ALL applications. These include accelerometers, gyroscopes, and magnetometers~\cite{Vincenzo,Bennasar,Khimraj}. Wearable devices are favoured for their mobility, portability, and accessibility. Numerous studies have utilised wearable devices to detect falls and alert caregivers~\cite{Kulurkar,Warrington}. However, this approach requires users to constantly wear the devices, which may cause discomfort. Forgetting to wear it can negate the purpose of monitoring, compromising the system's effectiveness. 

Non-wearable sensor solutions for HAR involve devices or systems capable of detecting and analysing human activities without direct attachment to the body. Examples include radio-frequency-based systems~\cite{Muaaz,Chao} and, increasingly, vision-based methods~\cite{Vishwakarma}. 

However, vision-based solutions have drawbacks, such as limited field of view, sensitivity to environmental factors like lightning conditions and cluttered backgrounds, and privacy concerns~\cite{Vishwakarma,Jegham}. They also suffer from subject occlusion, which occurs when parts of the subject's body are hidden or obscured by other objects or body parts within the room, leading to incomplete or inaccurate tracking of movements~\cite{Jegham}.

\subsection{Skeleton-based Action Recognition}\noindent
Skeleton-based action recognition approaches have gained attention owing to their lower computational complexity compared to processing RGB data. This has led to the development of methods aimed at enhancing action recognition performance.

Graph Convolutional Neural Networks (GCN) have become a fundamental framework in skeleton-based action recognition thanks to their ability to model the spatial and temporal dynamics of human joints effectively. One of the pioneering works is the Spatial-Temporal Graph Convolutional Network (ST-GCN) introduced in~\cite{yan2018spatial}, which captures spatial and temporal features by leveraging the graph structure of the skeleton data.

To enhance GCNs' representational ability, several methods have focused on adaptively learning the graph structure~\cite{shi2019two,bian2022self,si2019attention,shi2019skeleton,zhang2020semantics,zhang2020context,peng2020learning,li2021symbiotic,li2019actional,chen2021channel}. These methods employ adaptive GCNs that dynamically learn the graph's topology. 

Other approaches leverage attention mechanisms and transformer, and have integrated them into GCN frameworks to improve their ability to capture significant relationships~\cite{chi2022infogcn,shi2020decoupled,plizzari2021spatial}. These methods utilise attention mechanisms within graph convolutions to identify key relationships in the data.

Alternative approaches aim to enhance GCN performance by refining the input representation to make it more informative~\cite{li2019actional,shi2019skeleton,shi2019two}. These works define inputs in terms of bones and motion by preprocessing existing joint data. These methods aim to deepen the network's understanding of underlying action dynamics by providing richer features.

Despite these advancements, few methods address the practical constraints of deploying these systems in real-world environments. Many existing works are unsuitable for real-world deployment, particularly given the computational limitations of edge devices and the cluttered environments in real-world settings that introduce noise and occlusions into the data.

To address some of the real-world challenges, some approaches have concentrated on designing efficient architectures to improve real-time performance~\cite{song2020stronger,lin2019tsm,cheng2020skeleton,yang2021feedback,song2022constructing}.
Others have focused on mitigating the effects of noise and occlusion in real-world environments by proposing methods to improve performance under these challenging conditions~\cite{yoon2022predictively,Tabkhi,Zhenjie,Liang,guo2024overcomplete}. 

This paper aligns with these efforts by developing an efficient architecture that satisfies the real-time requirements of devices with limited computational resources. Additionally, we address the challenges posed by noise and occlusion in real-world environments to achieve robust performance in practical applications. Notably, our approach aims to achieve these objectives without sacrificing accuracy.  

\section{Method}
\label{s:Method}
\begin{figure*}[!t]
  \centering  \includegraphics[width=1\linewidth]{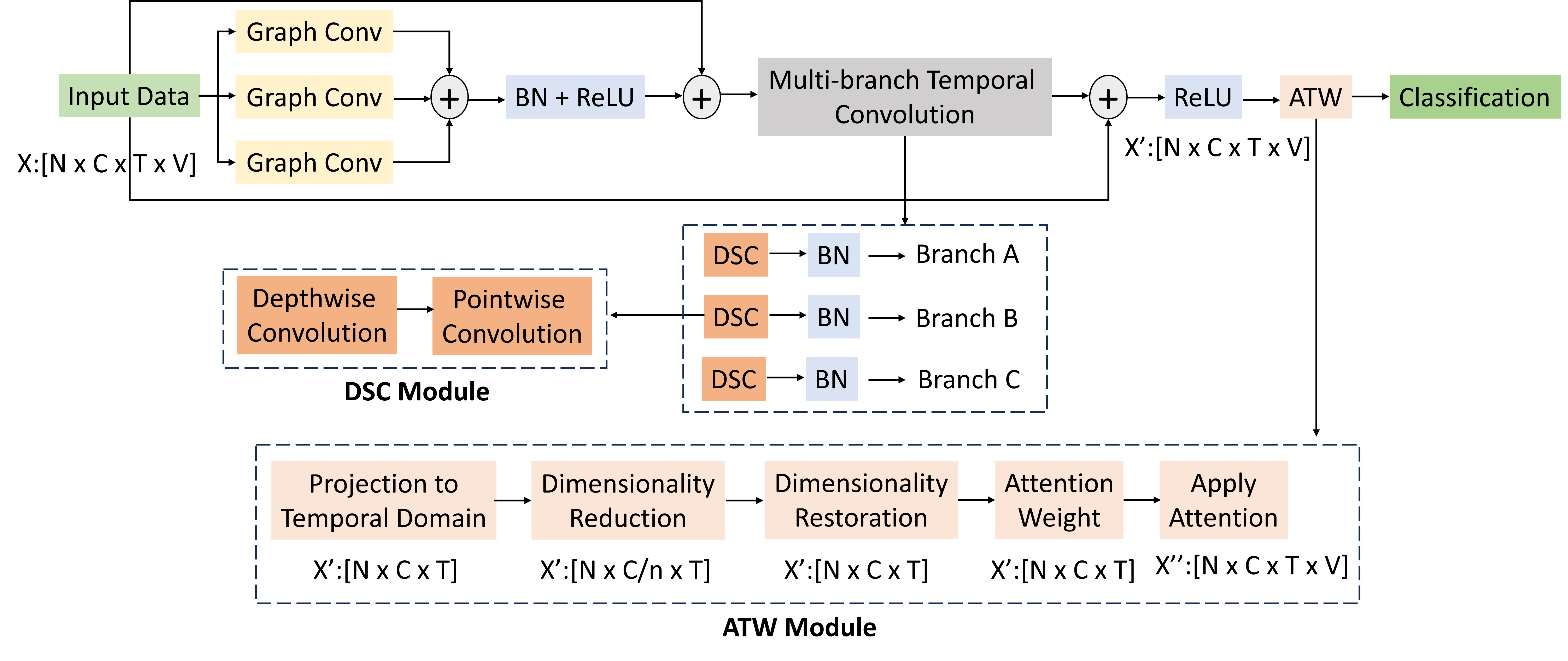}
  \caption{Architecture of the proposed RE-TCN: Graph convolution is first applied to the skeleton sequences. The output is then passed to the multi-branch temporal convolution, followed by the ATW mechanism, and finally to the classification module for action recognition. DSC and ATW denote Depthwise Separable Convolution and Adaptive Temporal Weighting, respectively.}
  \label{fig:method_figure}
\end{figure*}

\noindent
The architecture of RE-TCN is depicted in Fig.~\ref{fig:method_figure}. The model aims to enhance accuracy, robustness to noise and occlusion, and computational efficiency. The contributions of the method are threefold: an adaptive temporal weighting mechanism, depthwise separable convolution and data augmentation. Details of the implementation are explained below. 
The objective of human action recognition is to learn a function $f: X \rightarrow y$ that maps the spatio-temporal skeletal data $X$ to an action label $y$ by capturing spatial dependencies between joints and temporal progression across frames. 
Input data is denoted by $X$, with dimensions $N, C, T, V$, where $N$ is the batch size, $C$ is the number of channels (features), $T$ is the number of frames, and $V$ is the number of joints. 

\subsection{Adaptive Temporal Weighting}
\noindent
The Adaptive Temporal Weighting (ATW) mechanism is developed to dynamically assign different levels of importance to frames within an action sequence. This section outlines the implementation and the design decisions made during its development to optimise computational efficiency. 

\subsubsection{Implementation of ATW}
\noindent
ATW compute an attention weight, highlighting the importance of different frames for each input sequence.
To compute these weights, ATW first collapses the joint dimension by computing the average over joints.
\begin{equation}\label{eq:compute_average}
    X_{\text{proj}} = \frac{1}{V} \sum_{v=1}^{V} X[:,:,:,:v] 
\end{equation}
where $X_{\text{proj}} \in \mathbb{R}^{N \times C \times T}$. This step reduces the spatial complexity and focuses on the temporal aspect of the features. 

To efficiently compute weight, the temporal feature map is projected into lower-dimensional space using a $1\times1$ convolution,
\begin{equation}\label{eq:feature_projection}
    X_{\text{red}} = \text{Conv1}(X_{\text{proj}})
\end{equation}
where $X_{\text{red}} \in \mathbb{R}^{N \times \frac{C}{n} \times T}$. The convolution operation in this step is designed to serve two purposes. First, it reduces the number of channels by a factor of $n$, enabling more efficient computation while preserving temporal information. Secondly, the convolution operation acts as a feature transformation by combining the information from the temporal features within each channel. 

The reduced temporal feature map is restored to its original number of channels using another $1\times1$ convolution,
\begin{equation}\label{eq:feature_restoration}
    X_{\text{restored}} = \text{Conv2}(X_{\text{red}})
\end{equation}
where $X_{\text{restored}} \in \mathbb{R}^{N \times C \times T}$. This step further transforms the intermediate features and restores the dimensionality of the temporal feature map from $ \frac{C}{n}$ back to $C$, ensuring that the feature representation aligns with the original number of channels before weight is applied.

Next, a softmax function is applied across the temporal dimension to compute the attention weights. The softmax operation ensures that the assigned weights sum up to 1. 
\begin{equation}\label{eq:softmax}
    \alpha_t = \frac{\exp(X_{\text{restored}, t})}{\sum_{t=1}^{T} \exp(X_{\text{restored},t})}
\end{equation}
where $\alpha_t \in \mathbb{R}^{N \times C \times T}$ are the learned attention weights. 

Finally, the original input tensor is weighted by the attention scores along the temporal dimension: 
\begin{equation}\label{eq:apply_attention}
    X' = X \cdot \alpha_t
\end{equation}
where $X' \in \mathbb{R}^{N \times C \times T \times V}$ is the output of the ATW. Here, the learned attention weights are applied across the temporal dimension, scaling each frame according to its relative importance.  

In comparison to the TD-GCN~\cite{Xinshun}, the convolution operation applies convolutions across the temporal dimension as follows: 
\begin{equation}\label{eq:TD_GCN_Convolution}
    X_{\text{conv}} = \text{Conv}(X)
\end{equation}
where each frame is processed uniformly, leading to potential information loss in important frames. 
In the proposed approach, the temporal convolutions are complemented by ATW which assigns higher importance to informative frames. The difference lies in the introduction of the ATW mechanism that dynamically scales each frame. 
\begin{equation}\label{eq:ATW_Convolution}
    X_{\text{att}} = X \cdot \alpha_t
\end{equation}
This ensures that the model learns to focus on the most relevant temporal features, enhancing its ability to capture crucial moments in the action sequence. 

\subsubsection{Design Choices in ATW}
\noindent
The ATW mechanism is designed to achieve computational efficiency while maintaining modelling capacity. Inspired by the depthwise separable convolution process, which breaks down convolution into a two-step operation, ATW's core design replaces a single convolution with ($C_{\text{in}} \to C_{\text{out}}$) channels with two-step convolutions featuring an intermediate dimensionality reduction:
($C_{\text{in}} \to C_{\text{mid}} \to C_{\text{out}}$). This approach divides the convolution operation into two steps. The first convolution operation $C_{\text{in}} \to C_{\text{mid}}$ reduces the channels from $C_{\text{in}}$ to an intermediate $C_{\text{mid}}$ where $C_{\text{mid}} < C_{\text{in}}$, while the second convolution operation restores the channels back from $C_{\text{mid}}$ to $C_{\text{out}}$, where $C_{\text{out}}=C_{\text{in}}$.

The reduction in computational cost is significant. For instance, the computational cost for a $1\times1$ convolution operation can be expressed as:
\begin{equation}\label{eq:1_1_Convolution}
    X_{\text{Cost}} = N \times C_{\text{out}} \times C_{\text{in}} \times T \times V
\end{equation}
The cost for the first and second convolution operations can therefore be expressed as:
\begin{equation}\label{eq:ATW_Convolution_1}
    \text{Cost}_{\text{first}} = N \times C_{\text{mid}} \times C_{\text{in}} \times T \times V
\end{equation}
\begin{equation}\label{eq:ATW_Convolution_2}
    \text{Cost}_{\text{second}} = N \times C_{\text{out}} \times C_{\text{mid}} \times T \times V
\end{equation}
The total computational cost for this two-step process is:
\begin{align}\label{eq:ATW_Convolution_total}
    \text{Cost}_{\text{total}} = & N \times C_{\text{mid}} \times C_{\text{in}} \times T \times V \notag \\ & + N \times C_{\text{out}} \times C_{\text{mid}} \times T \times V
\end{align}
For comparison, a single convolution operation from $C_{\text{in}} \to C_{\text{out}}$ requires:
\begin{equation}\label{eq:Convolution_single}
    \text{Cost}_{\text{single}} = N \times C_{\text{out}} \times C_{\text{in}} \times T \times V
\end{equation}
Using two convolutions with an intermediate reduction results in evident computational savings. The ratio of computational cost between the two-step process and the single convolution is: 
\begin{equation}\label{eq:ratio}
    \frac{\text{Cost}_{\text{total}}}{\text{Cost}_{\text{single}}} = \frac{C_{\text{mid}} \times C_{\text{in}} + C_{\text{out}} \times C_{\text{mid}}}{C_{\text{out}} \times C_{\text{in}}}
\end{equation}
For $C_{\text{mid}} \ll C_{\text{in}}$, this ratio becomes much smaller, demonstrating that the two-step design is significantly more efficient. 

\subsection{Depthwise Separable Convolutions}
\noindent
TD-GCN~\cite{Xinshun} employs convolutions across both temporal and spatial dimensions, enabling it to capture spatio-temporal features effectively and thereby improving the understanding of temporal dynamics and spatial relationships between joints. Despite its strengths, the convolution operation in TD-GCN is computationally intensive and demands a relatively high number of parameters. To reduce the computational cost and improve the model efficiency without sacrificing accuracy, the proposed approach decomposes the convolution operation of TD-GCN into two stages; depthwise and pointwise convolution. The depthwise convolution performs a convolution operation for each input channel (feature), allowing for spatial feature extraction independently on each channel. In contrast, the pointwise convolution uses a $1\times1$ kernel to combine the outputs of the depthwise convolution across channels.

\subsubsection{Convolution Operation Process}
\noindent
Given an input data $X \in \mathbb{R}^{N \times C_{\text{in}} \times T \times V}$, where $N$ is the batch size, $C_{\text{in}}$ is 
the number of input channels, $T$ is the number of frames, and $V$ is the number of joints, TD-GCN~\cite{Xinshun} applies a set of filters $W \in \mathbb{R}^{C_{\text{out}} \times C_{\text{in}} \times K_h \times K_w}$, where $C_{\text{out}}$ is the number of output channels and $K_h \times K_w$ is the filter size. The output of the convolution operation is computed as
\begin{equation}\label{eq:TD_GCN_Conv}
    Y_{\text{TD-GCN}} = W \ast X,
\end{equation}
where $Y_{\text{TD-GCN}} \in \mathbb{R}^{N \times C_{\text{out}} \times T \times V}$ is the result of the convolution operation, which involves $C_{\text{in}} \times C_{\text{out}} \times K_h \times K_w$ multiplication. This approach requires a large number of parameters and high computational cost, especially when the input data is large. 

The proposed convolution operation is formulated as follows: 
\begin{equation}\label{eq:DW_Conv}
    Y_{\text{dw}} = W_{\text{dw}} \ast X
\end{equation}
\begin{equation}\label{eq:PW_Conv}
    Y_{\text{final}} = W_{\text{pw}} \ast Y_{\text{dw}}
\end{equation}
where $ W_{\text{dw}} \in \mathbb{R}^{C_{\text{in}} \times K_h \times K_w} $ is the depthwise filter applied to each input channel individually, $ W_{\text{pw}} \in \mathbb{R}^{C_{\text{out}} \times C_{\text{in}} \times 1 \times 1} $ is the pointwise filter, which aggregates the output of the depthwise convolution across channels, $ Y_{\text{dw}} \in \mathbb{R}^{N \times C_{\text{in}} \times T \times V} $ is the intermediate result after the depthwise convolution and $ Y_{\text{final}} \in \mathbb{R}^{N \times C_{\text{out}} \times T \times V} $ is the final output after the pointwise convolution. 

\subsubsection{Efficiency Comparison}
\noindent
For TD-GCN~\cite{Xinshun} convolution operation, the computational complexity can be expressed as: 
\begin{equation}\label{eq:TD_GCN_Conv_cost}
    \text{Cost}_{\text{TD-GCN}} = T \times V \times C_{\text{in}} \times C_{\text{out}} \times K_h \times K_w
\end{equation}
This requires $C_{\text{in}} \times C_{\text{out}} \times K_h \times K_w$ multiplications for each spatial location in the input, resulting in a significant computational cost when $C_{\text{in}}$ and $C_{\text{out}}$ are large. 

In contrast, the computational cost of depthwise separable convolution can be divided into depthwise and pointwise convolution. In depthwise convolution, each input channel is convolved independently with a single filter, leading to a computational cost of: 
\begin{equation}\label{eq:DW_Conv_cost}
    \text{Cost}_{\text{dw}} = T \times V \times C_{\text{in}} \times \times K_h \times K_w
\end{equation}
While in pointwise convolution, a $1\times1$ convolution is applied across all input channels, resulting in a computational cost of:
\begin{equation}\label{eq:PW_Conv_cost}
    \text{Cost}_{\text{pw}} = T \times V \times C_{\text{in}} \times C_{\text{out}} 
\end{equation}
As a result, the total cost of the depthwise separable convolution is: 
\begin{align}\label{eq:DW_PW_Conv_cost}
    \text{Cost}_{\text{DSC}} = & T \times V \times C_{\text{in}} \times K_h \times K_w \notag \\ & + C_{\text{in}} \times C_{\text{out}}
\end{align}
Comparing the computational costs of the two approaches, we can see that the depthwise separable convolution is more efficient:
\begin{equation}\label{eq:Conv_cost_compare}
    \frac{\text{Cost}_{\text{DSC}}}{\text{Cost}_{\text{TD-GCN}}} = \frac{C_{\text{in}} \times K_h \times K_w + C_{\text{in}} \times C_{\text{out}}}{ C_{\text{in}} \times C_{\text{out}} \times K_h \times K_w}
\end{equation}
For a large $K_h$ and $K_w$, the depthwise separable convolution offers a substantial reduction in computation, especially when $C_{\text{in}}$ and $C_{\text{out}}$ are large. 

\subsection{Data Augmentation}
\begin{figure*}[!tb]
  \centering  \includegraphics[width=1\linewidth]{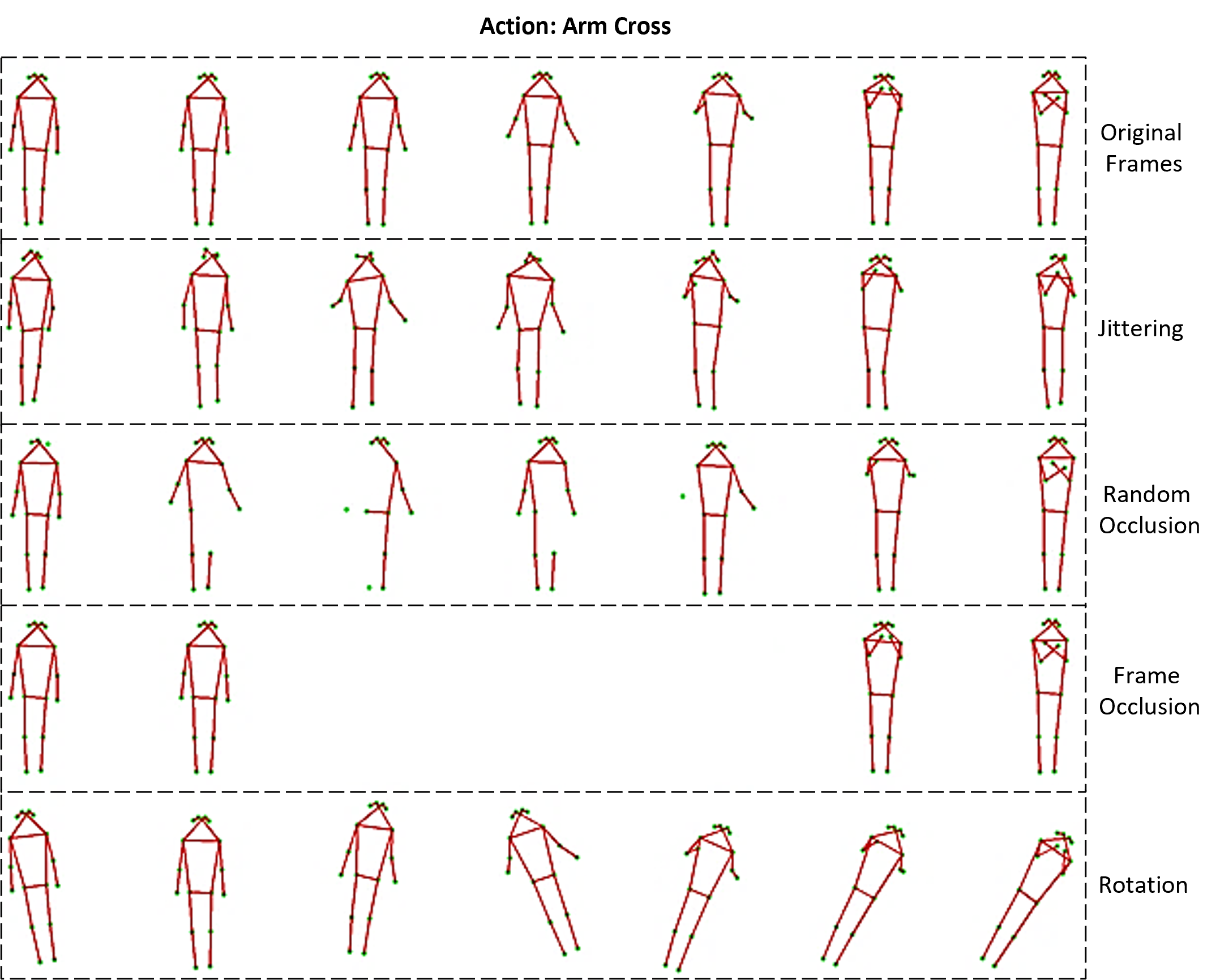}
  \caption{A skeleton sample of "Cross Arm" action with data augmentation techniques: jittering, random occlusion, frame occlusion, and rotation }
  \label{fig:data_aug}
\end{figure*}

% \noindent
To enhance robustness against noise and occlusion in real-world environments, we propose a suite of augmentation techniques (see Fig.~\ref{fig:data_aug}) to model conditions that lead to performance degradation. For instance, we recognised that an object occluding a camera view typically affects a continuous sequence of frames in real-world scenarios. This can occur randomly due to the cluttered real-world environments often seen. With these practical scenarios in mind, the following augmentation strategies are designed.

\subsubsection{Random Joint and Frame Occlusion}
% \noindent
This method introduces variability into the spatial and temporal dimensions of the skeleton data, simulating potential occlusion situations encountered in real-world environments. It applies random erasure of joints and frames across randomly selected continuous sequences of frames. 
The process is controlled by the probability of erasure $p$ and erasing sequences length range $L_{\text{min}}$ and $L_{\text{max}}$. First, frames are chosen for erasure based on probability $p$. Consecutive sequences of frames are then selected, with lengths between $L_{\text{min}}$ and $L_{\text{max}}$. For each selected sequence, random joints are set to zero to simulate joint occlusion, while for frame occlusion, all joint values are set to zero. After processing each sequence, a random number of frames are skipped before the subsequent process begins. The full procedure is outlined in Algorithm~\ref{alg:occlusion}
\begin{algorithm}
\caption{Random Joint and Frame Occlusion}\label{alg:occlusion}
\begin{algorithmic}[1]
\Require Original skeleton data $d \in \mathbb{R}^{C \times T \times V \times M}$ 
\Ensure Augmented skeleton data with occlusions $d_{\text{aug}} \in \mathbb{R}^{C \times T \times V \times M}$
\State Initialize $d_{\text{aug}}$ with $d$
\State $C, T, V, M \leftarrow \text{Shape of } d$
\State $p, L_{\text{min}}, L_{\text{max}} \leftarrow \text{Probability, Min/Max occlusion length}$

\For{$m \gets 0$ to $M-1$}  
    \State $t_{\text{current}} \gets 0$ 
    \While{$t_{\text{current}} < T$}
        \State $t_{\text{remaining}} \gets T - t_{\text{current}}$
        \If{$t_{\text{remaining}} < L_{\text{min}}$}
            \State \textbf{break}
        \EndIf

        \State $L_{\text{occlusion}} \gets \text{Random int in } [L_{\text{min}}, \min(L_{\text{max}}, t_{\text{remaining}})]$
        
        \If{Joint Occlusion is selected} 
            \For{$t \gets t_{\text{current}}$ to $t_{\text{current}} + L_{\text{occlusion}} - 1$}
                \State $J_{\text{occlude}} \gets \text{Random subset of joints from } \{1, 2, \dots, V\}$
                \State Set $d_{\text{aug}}[:, t, J_{\text{occlude}}, m] \gets 0$
            \EndFor
        \ElsIf{Frame Occlusion is selected}
            \State Set $d_{\text{aug}}[:, t_{\text{current}} : t_{\text{current}} + L_{\text{occlusion}} - 1, :, m] \gets 0$
        \EndIf

        \State $t_{\text{current}} \gets t_{\text{current}} + L_{\text{occlusion}}$
        
        \State $t_{\text{skip}} \gets \text{Random int in } [1, \min(10, T - t_{\text{current}})]$
        \State $t_{\text{current}} \gets t_{\text{current}} + t_{\text{skip}}$
    \EndWhile
\EndFor
\end{algorithmic}
\end{algorithm}
\subsubsection{Skeleton Rotation}
\noindent
This method introduces variability in spatial orientations to simulate different viewing angles. First, a random rotation vector is generated and transformed into a rotation matrix. The rotation matrix is then multiplied with the skeleton data to rotate the joint positions. The full algorithm is presented in algorithm~\ref{alg:random_rot}
\begin{algorithm}
\caption{Random Rotation of Skeleton Data}\label{alg:random_rot}
\begin{algorithmic}[1]
\Require Skeleton data $d \in \mathbb{R}^{C \times T \times V \times M}$, rotation angle $\theta$
\Ensure Augmented skeleton data with random rotation $d_{\text{rot}} \in \mathbb{R}^{C \times T \times V \times M}$
\State Convert $d$ to Torch tensor $d_{\text{torch}}$
\State $C, T, V, M \leftarrow \text{Shape of } d_{\text{torch}}$
\State Reshape and permute $d_{\text{torch}} \leftarrow \text{Reshape to } (T, C, V \times M)$
\State Initialize random rotation vector $\text{rot} \in \mathbb{R}^{T \times 3}$ with values in $[-\theta, \theta]$
\State Apply a rotation function $\_rot$ to create a rotation matrix $\text{rot} \gets \text{\_rot}(\text{rot}) \in \mathbb{R}^{T \times 3 \times 3}$
\State Multiply rotation matrix with skeleton data $d_{\text{torch}} \leftarrow \text{rot} \times d_{\text{torch}}$
\State Reshape and permute $d_{\text{torch}} \leftarrow \text{Reshape back to } (C, T, V, M)$
\State Return augmented skeleton data $d_{\text{rot}}$
\end{algorithmic}
\end{algorithm}
% \begin{algorithm}
%     \caption{Skeleton Rotation} \label{alg:random_rot}
%     \begin{algorithmic}[1]
%         \Require Skeleton data $d \in \mathbb{R}^{C \times T \times V \times M}$, rotation angle $\theta$
%         \Ensure Augmented skeleton data with random rotation $d_{\text{rot}} \in \mathbb{R}^{C \times T \times V \times M}$
%         \State Convert $d$ to Torch tensor $d_{\text{torch}}$
%         \State $C, T, V, M \leftarrow \text{Shape of } d_{\text{torch}}$
%         \State Reshape and permute $d_{\text{torch}} \leftarrow \text{Reshape to } (T, C, V \times M)$
%         \State Initialise random rotation vector $\text{rot} \in \mathbb{R}^{T \times 3}$ with values in $[-\theta, \theta]$
%         \State Apply the rotation to create a rotation matrix $\text{rot} \gets \text{\_rot}(\text{rot}) \in \mathbb{R}^{T \times 3 \times 3}$
%         \State Multiply rotation matrix skeleton data $d_{\text{torch}} \leftarrow \text{rot} \times d_{\text{torch}}$
%         \State Reshape and permute $d_{\text{torch}} \leftarrow \text{Reshape back to } (C, T, V, M)$
%         \State Return augmented skeleton data $d_{\text{rot}}$
%     \end{algorithmic}
% \end{algorithm}
\subsubsection{Jittering}
\noindent
The jittering method introduces random perturbations to skeleton data by adding Gaussian noise to the joint position in selected frames. First, frames are chosen for noise addition based on a random decision controlled by the probability $p_\text{frame}$. For each selected frame, Gaussian noise is generated and added to the joint positions. The algorithm describing this process is detailed in Algorithm~\ref{alg:jittering}.
\begin{algorithm}
    \caption{Jittering}\label{alg:jittering}
    \begin{algorithmic}[1]
        \Require Skeleton data $d \in \mathbb{R}^{C \times T \times V \times M}$, Gaussian noise standard deviation $\sigma$, frame selection probability $p_{\text{frame}}$
        \Ensure Augmented skeleton data with jittering $d_{\text{jittered}} \in \mathbb{R}^{C \times T \times V \times M}$
        \State Initialise augmented data $d_{\text{jittered}} \gets \text{copy of } d$
        \State $C, T, V, M \leftarrow \text{Shape of } d$
        \For{$m \gets 0$ to $M-1$}
            \For{$t \gets 0$ to $T-1$}
                \If{$\text{rand}() < p_{\text{frame}}$}
                    \State Generate Gaussian noise $\text{noise} \sim \mathcal{N} (0, \sigma^2)$ of shape $(C \times V)$
                    \State Add noise to the selected frame: $d_{\text{jittered}}[:, t, :, m] \gets d_{\text{jittered}}[:, t, :, m] + \text{noise}$
                \EndIf
            \EndFor
        \EndFor
        \State Return augmented skeleton data $d_{\text{jittered}}$
    \end{algorithmic}
\end{algorithm}

\section{Experiments}
\label{s:Experimental}
\noindent
In this section, we evaluate the accuracy, robustness, and efficiency of the proposed RE-TCN framework. We compare RE-TCN's performance to state-of-the-art skeleton-based action recognition methods and conduct comprehensive ablative studies.

\subsection{Dataset}
\noindent
1) \textbf{NTU RGB+D 60}~\cite{7780484}: The dataset comprises 56,880 skeleton sequences covering 60 actions performed by 40 subjects. The authors~\cite{7780484} suggest two primary evaluation benchmarks: (1) \textbf{Cross-View (X-view)}, where training data is captured from two camera angles, $0^{\circ}$ (view 2) and $45^{\circ}$ (view 3), while testing is conducted from $-45^{\circ}$ (view 1), and (2) \textbf{Cross-Subject (X-sub)}, in which 20 subjects data is used for training while data from the remaining 20 subjects is reserved for testing. Following this approach, we report the top-1 recognition accuracy across both benchmarks. 

\noindent
2) \textbf{Northwestern-UCLA (NW-UCLA)}~\cite{wang2014crossview}: The NW-UCLA dataset includes 1,494 sequences across 10 actions, captured from Kinect cameras, each positioned to provide different viewpoints. Following the evaluation protocol suggested by the authors~\cite{wang2014crossview}, data from the first two cameras is used for training, and the remaining camera's data is used for testing. 
\noindent

3) \textbf{SHREC'17}~\cite{LI20151}: The dataset comprises 2,800 gesture sequences performed by 28 participants. Each of the 28 participants performed each gesture 10 times. The gestures are categorised into either 14 or 28 classes based on the gesture type. Consistent with the evaluation protocol outlined in~\cite{LI20151}, 1,960 sequences were used for training, and 840 were reserved for testing. 
 \noindent

4) \textbf{DHG-14/28}~\cite{de2016skeleton}: The dataset comprises 2,800 gesture sequences performed 5 times each by 20 participants. As suggested by the authors~\cite{de2016skeleton}, the leave-one-subject-out cross-validation method is used for evaluation. This means that data from 19 participants is used for training and the remaining participant's data is reserved for testing. This evaluation process is repeated 20 times, and the final accuracy is reported as the average of these iterations.
 
\subsection{Implementation Details}
The model was trained using Stochastic Gradient Descent (SGD) with a warm-up strategy. The learning rate was initialised at 0.1, and a momentum of 0.9 was applied. Model checkpointing was used to save the model based on optimal performance on the validation set. For the SHREC'17 and DHG-14/28 datasets, a batch size of 32 was used, with a weight decay of 0.0001 and a learning rate decay factor of 0.1. The NTU RGB+D 60 dataset was trained with a batch size of 64, a weight decay of 0.0004 and a learning rate decay factor of 0.1. For the NW-UCLA dataset, the model was trained with a batch size of 16, a weight decay of 0.0001 and a learning rate decay factor of 0.1. Based upon findings from ablation studies, the following parameter settings were used for the ATW mechanism: the reduction ratio was set to 8, the mean pooling strategy was used, placement in the model architecture was set to late, and two layers with a reduction ratio were implemented.

\subsection{Ablation Study and Parameter Tunning}
To analyse the various components of the proposed method, we performed extensive experiments using the NW-UCLA dataset as a case study.

\subsubsection{Adaptive Temporal Weighting (ATW) and Depthwise Separable Convolution (DSC)}
We evaluate the effectiveness of the proposed ATW and DSC in enhancing model accuracy and optimising parameter count, as summarised in Tab~\ref{tab:ATW_DSC_impact}. For this evaluation, baseline training was conducted using the TD-GCN~\cite{Xinshun} and TD-GDSCN~\cite{Gbouna} model, and the same training parameter settings were maintained across all experiments. The only modifications involved the integration of different components of the proposed method. As shown in Tab~\ref{tab:ATW_DSC_impact}, the integration of DSC improved both accuracy and parameter efficiency. Furthermore, the addition of ATW yielded an additional increase in accuracy, although with a minimal parameter increase of 0.01. These results validate the contribution of DSC and ATW in enhancing the baseline model's performance. 
\begin{table}[htb]
  \caption{Impact of Adaptive Temporal Weighting (\textbf{ATW}) and Depthwise Separable Convolution (\textbf{DSC}) on Accuracy: The best result is highlighted in bold.}
  \vspace{10pt}
  \label{tab:ATW_DSC_impact}
  \centering
  \footnotesize
  \begin{tabular}{@{}llll@{}}
    \toprule
    {Method} & {Parameters (M)} & {Accuracy (\%)} \\
    \midrule
    {TD-GCN} & 1.35 & 94.40 \\
    {TD-GDSCN}  & \textbf{1.23} & 95.05\\
    {RE-TCN + DSC}  & \textbf{1.23} & 95.69\\
    {RE-TCN + DSC + ATW}  & 1.24 & \textbf{96.34} \\
  \bottomrule
  \end{tabular}
\end{table}

\subsubsection{Reduction Ratio}
To enable efficient convolution operations, one design choice in ATW was to reduce the number of channels by a factor of $n$ in the first convolution operation. We conducted an experiment to select an appropriate reduction ratio. Initially, we used a reduction ratio of 8 to train the model as a baseline. We then tested different reduction ratios and presented the results in Tab~\ref{tab:reduction_ratio_impact}. The results show that higher reduction ratios lead to fewer parameter counts, with a ratio of 64 yielding the fewest parameters. However, accuracy fluctuated among the different ratios, with the ratio of 8 achieving the highest accuracy. Since the increase in parameter count with a ratio of 8 compared to 64 was minimal, and the accuracy improvement was significant, we used the reduction ratio of 8 in subsequent experiments.  
\begin{table}[htb]
  \caption{Impact of Reduction Ration on Accuracy: The best result is highlighted in bold}
  \vspace{10pt}
  \label{tab:reduction_ratio_impact}
  \centering
  \footnotesize
  \begin{tabular}{@{}llll@{}}
    \toprule
    {Reduction Ratio} & {Accuracy (\%)} & {Parameters} \\
    \midrule
    4 & 93.32 & 1258896 \\
    8 & \textbf{96.34} & 1242480 \\
    16 & 92.24 & 1234272\\
    32 & 94.40 & 1230168\\
    64 & 93.75 & \textbf{1228116}\\
  \bottomrule
  \end{tabular}
\end{table}

\subsubsection{Pooling Strategy}
To compute the attention weights along the temporal dimension, the ATW mechanism first collapses the joint dimension. This reduces the spatial complexity and focuses on the temporal aspect of the features. We conducted an experiment to determine the most appropriate pooling mechanism for collapsing the joint dimension. We tested various pooling strategies and presented the results in Tab~\ref{tab:pooling_impact}. The results showed that all pooling strategies had the same effect on parameter count. However, the mean strategy's accuracy was significantly higher compared to others. Therefore, we selected it as the pooling strategy for subsequent experiments. 
\begin{table}[htb]
  \caption{Impact of Pooling Strategy on Accuracy: The best result is highlighted in bold}
  \vspace{10pt}
  \label{tab:pooling_impact}
  \centering
  \footnotesize
  \begin{tabular}{@{}llll@{}}
    \toprule
    {Pooling Strategy} & {Accuracy (\%)} & {Parameters (M)} \\
    \midrule
    Max & 93.10 & 1.24 \\
    Adaptive & 94.18 & 1.24 \\
    Mean & \textbf{96.34} & 1.24 \\
    Global Max & 91.81 & 1.24 \\
    Global Mean &  83.41 & 1.24 \\
  \bottomrule
  \end{tabular}
\end{table}

\subsubsection{Adaptive Temporal Weighting Location in Model Architecture}
We conducted an experiment to identify the optimal placement location of the ATW mechanism within the overall model architecture. Various positions were tested, with the result presented in Tab~\ref{tab:ATW_Location_impact}. The results indicate that the placement of ATW affects both accuracy and parameter count. Introducing the ATW mechanism earlier in the network minimised parameter count, though this configuration did not achieve the highest accuracy. Conversely, positioning ATW later in the network produced the highest accuracy, with only a minimal increase in parameter count relative to early placement. 
\begin{table}[htb]
  \caption{Impact of Adaptive Temporal Weighting (\textbf{ATW}) Location in Model Architecture on Accuracy: The best result is highlighted in bold}
  \vspace{10pt}
  \label{tab:ATW_Location_impact}
  \centering
  \footnotesize
  \begin{tabular}{@{}llll@{}}
    \toprule
    {Location in Model Architecture} & {Accuracy (\%)} & {Parameters} \\
    \midrule
    Early & 93.10 & \textbf{1226904}\\
    Middle & 92.75 & 1230048\\
    Late & \textbf{96.34} & 1242480 \\
    Early + Middle & 92.89 & 1231144\\
    Late + Middle & 92.67 & 1246720\\
    Early + Middle + Late & 93.10 & 1247816\\
  \bottomrule
  \end{tabular}
\end{table}

\subsubsection{Efficient Design Strategy}
One key design choice in ATW mechanism was to use a two-step convolution operation with an intermediate dimensionality reduction. We conducted an experiment using various combinations of convolution layers to assess their impact on accuracy and parameter count. The result is presented in Tab~\ref{tab:ATW_design_strategy_impact}. We trained and evaluated two sets of models: one with only full convolution layers (without a dimensionality reduction) and another with a dimensionality reduction layer. In both sets, we observed that the parameter count increased with the number of convolution layers. However, accuracy did not consistently improve with more layers. In both sets, the accuracy was higher with two convolution layers, with the dimensionality reduction model performing best. Based on this finding, we adopted this setup for subsequent experiments. 
\begin{table}[!htb]
  \caption{Impact of Design Strategy on Accuracy and Parameter Count: Layer = Convolution layers, W-red = With reduction ration of 8 and the best result is highlighted in bold}
  \vspace{10pt}
  \label{tab:ATW_design_strategy_impact}
  \centering
  \footnotesize
  \begin{tabular}{@{}llll@{}}
    \toprule
    {Design Strategy} & {Accuracy (\%)} & {Parameters} \\
    \midrule
    One layer & 93.97 & 1291600\\
    Two layers & 94.40 & 1357392 \\
    Three layers & 81.41 & 1423184 \\
    Four layers & 88.58 & 1488976\\
    Two layers/W-red & \textbf{96.34} & \textbf{1242480} \\
    Three layers/W-red & 89.44 & 1308272 \\
    Four layers/W-red & 91.51 & 1374064 \\
  \bottomrule
  \end{tabular}
\end{table}

\subsubsection{Data Augmentation Strategy}
We explored the accuracy of different data augmentation strategies to evaluate their effect. The result is presented in Tab~\ref{tab:table_data_augmentation}. First, we trained the model without data augmentation and tested it on data with jittering and occlusion. Next, we introduced various augmentation types and tested them again on data with jittering and occlusion. The results show that accuracy increases significantly with the introduction of data augmentation. This validates the effectiveness of data augmentation in ensuring robust action recognition against noise. 
\begin{table}[htb]
  \caption{Influence of Augmentation Types: N = Noise (jittering and occlusion), and R = Skeleton rotation: The best result is highlighted in bold}
  \vspace{10pt}
  \label{tab:table_data_augmentation}
  \centering
  \footnotesize	
  \begin{tabular}{@{}lll@{}}
    \toprule
    {Augmentation type} & {Jittering (\%) } & {Occlusion (\%) } \\
    \midrule
    RE-TCN & 88.36 & 82.54 \\
    RE-TCN+R & 92.24 & 90.73 \\
    RE-TCN+R+N & \textbf{94.18} & \textbf{94.40} \\
  \bottomrule
  \end{tabular}
\end{table}

\subsection{Performance Comparison with TD-GCN}
\subsubsection{Comparison of Accuracy and Parameter Count}
We conducted an experiment using the SHREC'17 and NW-UCLA datasets to compare the accuracy and parameter count of RE-TCN and TD-GCN. For a fair comparison, we used the same training parameter settings and joint data modality for both models. The result is presented in Tab~\ref{tab:table_model_computation_cost_comparison}. RT-TCN outperformed TD-GCN across both datasets, with accuracy improvements of 1.54\% for NW-UCLA, 3.54\% for SHREC'17 14 gesture, and 6.38\% for SHREC'17 28 gesture. In terms of computational efficiency, RE-TCN reduced the parameter count by 0.11\% for both datasets. These findings demonstrate that RE-TCN effectively enhances both computational efficiency and accuracy. 

\begin{table}[htb]
  \caption{Comparison of Accuracy and Parameter Count: The best result is highlighted in bold}
  \vspace{10pt}
  \label{tab:table_model_computation_cost_comparison}
  \centering
  \footnotesize
  \begin{tabular}{@{}lllll@{}}
    \toprule
    {Method} & {Dataset} & {Parameters (M)} & {Accuracy (\%)}\\
    \midrule
    TD-GCN & NW-UCL & 1.35 & 94.8\\
    RE-TCN (Ours) & NW-UCL & \textbf{1.24} & \textbf{96.34} \\ % \hdashline
    \hline
    TD-GCN & SHREC'17 14 & 1.36 & 96.31 \\
    RE-TCN (Ours) & SHREC'17 14 & \textbf{1.25} & \textbf{99.85} \\ %\hdashline
    \hline
    TD-GCN & SHREC'17 28 & 1.36 & 93.57 \\
    RE-TCN (Ours) & SHREC'17 28 & \textbf{1.25} & \textbf{99.95} \\ 
  \bottomrule
  \end{tabular}
\end{table}

\subsubsection{Performance Comparison on Cross-subject Evaluation}
We conducted an experiment utilising the DHG-14/28 dataset to compare the generalisability capability of TD-GCN and RE-TCN. We aim to evaluate the ability of the model to generalise across new participants whose data were not used to train the model. We utilised the DHG-14/28 dataset as it allows effective cross-subject evaluation through the leave-one-subject-out cross-validation method suggested by the authors~\cite{de2016skeleton}. To ensure a fair comparison, we employed the same training parameter settings and joint data modality for both models. The result is presented in Tab~\ref{tab:dhg_14_28_dataset}. 

From the result, we observe that performance across the subjects varies, highlighting the differences that exist when individuals perform the same action. These variations in how actions are performed by different people make recognising actions more challenging for certain individuals, as evidenced by the varying accuracy amongst subjects. Nevertheless, despite this challenge, the overall accuracy of RE-TCN across all the subjects for both 14 and 28 action types was notably high. When compared to TD-GCN, the accuracy was comparable, with RE-TCN demonstrating superior performance in some subjects while achieving equal accuracy in others. Overall, the result underscores the capability of RE-TCN to generalise effectively across different subjects. 

\begin{table*}[htb]
  \caption{Accuracy Recognition per Subjects for the DHG-14/28 Dataset: The - sign denotes results not provided by TD-GCN and the results are in accuracy (\%) with the best result highlighted in bold}
  \vspace{10pt}
  \label{tab:dhg_14_28_dataset}
  \centering
  \footnotesize
  \setlength{\tabcolsep}{2.5pt}
  % \tiny
  \begin{scriptsize}
  \begin{tabular}{@{}lllllllllllllllllllll@{}}
    \toprule
    \textbf{Dataset Type} / \textbf{Subject} & {1} & {2} & {3} & {4}  & {5} & {6} & {7} & {8} & {9} & {10} & {11} & {12} & {13} & {14}  & {15} & {16} & {17} & {18} & {19} & {20}  \\
    \midrule
    14 Gestures (RE-TCN) (\%)  & 88.57 & 80.00 & 97.86 & \textbf{95.00} & \textbf{92.86} & 87.86 & 87.14 & 92.86 & 91.43 & \textbf{95.00} & 95.00 & \textbf{92.14} & 88.57 & \textbf{90.00} & 95.71 & 91.14 & \textbf{87.14} & 91.43 & 97.86 & 88.57\\
    14 Gestures (TD-GCN) (\%)  & \textbf{90.00} & - & \textbf{98.57} & 93.57 & 90.00 & - & - & 92.86 & 91.43 & 92.86 & \textbf{95.71} & 90.00 & - & 88.57 & \textbf{97.14} & \textbf{96.43} & 86.43 & - & 97.86 & -\\
    \hdashline
    28 Gestures (RE-TCN) (\%)  & \textbf{90.00} & 72.86 & 92.14 & 88.57 & \textbf{90.71} & 84.29 & 91.43 & 86.43 & \textbf{85.71} & 92.86 & \textbf{92.86} & 90.71 & 83.57 & 84.29 & 95.00 & \textbf{90.71} & 84.29 & 90.71 & 93.57 &  83.57 \\
    28 Gestures TD-GCN) (\%)  & 89.29 & - & \textbf{93.57} & 88.57 & 85.00 & - & - & \textbf{92.14} & 83.57 & 92.86 & 89.29 & \textbf{92.14} & - & \textbf{88.57} & \textbf{95.71} & 87.86 & \textbf{86.43} & - & \textbf{94.29} &  - \\
  \bottomrule
  \end{tabular}
  \end{scriptsize}
\end{table*}

\subsection{Comparison with State-of-the-Art Methods}
Here, we compare RE-TCN's accuracy and robustness against noise and occlusion with state-of-the-art methods using the SHREC'17, NW-UCLA, and  NTU-RGB+D 60 datasets. Details of the comparison are discussed below. 
\subsubsection{Accuracy Comparison with State-of-the-Art Methods}
We compare RE-TCN with the state-of-the-art methods on the SHREC'17 and NW-UCLA datasets, which are skeleton-based gesture and skeleton-based action recognition datasets. Some methods use an ensemble approach, fusing results from joint, bone, and motion modalities, while others use only joint data. For a fair comparison, we report accuracy using only joint data modality. The results are presented in Tab~\ref{tab:NW_UCLA_sota_Compare}, and~\ref{tab:shrec17_sota_compare}. 

On both datasets, our model (RE-TCN) outperforms existing methods. In Tab~\ref{tab:shrec17_sota_compare}, using 14 and 28 gesture classes, classification accuracy is 99.95\% and 99.85\% respectively. This surpasses the current best-performing method~\cite{shi2020decoupled} by 2.85\% for 14 gestures and~\cite{Kyeongbo} by 5.05\% for 28 gestures. In Tab~\ref{tab:NW_UCLA_sota_Compare}, the classification accuracy is 96.34\%, outperforming the current best-performing methods\cite{CongqiECCV,yang2021feedback} by 1.04\%. These results demonstrate the effectiveness of the proposed in enhancing accuracy for both gestures and action recognition. 
\begin{table}[!hbt]
  \caption{Accuracy Comparison with State-of-the-Art Methods using SHREC'17 Dataset: The best result is highlighted in bold.}
  \vspace{10pt}
  \label{tab:shrec17_sota_compare}
  \centering
  \footnotesize
  \begin{tabular}{@{}llll@{}}
    \toprule
    {Method} & {14 Gestures (\%)} & {28 Gestures (\%)} \\
    \midrule
    ST-GCN~\cite{yan2018spatial} & 92.7 & 87.7 \\
    ST-TS-HGR-Net~\cite{Nguyen2019ANN}  & 94.3 & 89.4 \\
    HPEV~\cite{Yongcheng}  & 94.9 & 92.3 \\
    DSTA-Net~\cite{shi2020decoupled}  & 97.0 & 93.9 \\
    MS-ISTGCN~\cite{Kyeongbo}  & 96.7 & 94.9 \\
    TD-GCN~\cite{Xinshun}  & 96.31 & 93.57 \\
    RE-TCN (Ours) & \textbf{99.85} & \textbf{99.95} \\
  \bottomrule
  \end{tabular}
\end{table}
\begin{table}[!htb]
  \caption{Accuracy Comparison with State-of-the-Art Methods using NW-UCLA Dataset: The best result is highlighted in bold}
  \vspace{10pt}
  \label{tab:NW_UCLA_sota_Compare}
  \centering
  \footnotesize
  \begin{tabular}{@{}lll@{}}
    \toprule
    {Method} & {Accuracy (\%)}\\
    \midrule
    Lie Group~\cite{Vivek} & 74.2 \\
    Actionlet ensemble~\cite{Wang2014LearningAE} & 76.0 \\
    HBRNN-L~\cite{Du} & 78.5 \\
    Skeleton Visualisation~\cite{Mengyuan} & 86.1 \\
    Ensemble TS-LSTM~\cite{lee2017ensemble} & 89.2 \\
    AGC-LSTM~\cite{si2019attention} & 93.3 \\
    Shift-GCN~\cite{cheng2020skeleton} &  94.6\\
    DC-GCN+ADG~\cite{CongqiECCV} & 95.3 \\
    FGCN~\cite{yang2021feedback} &  95.3 \\
    TD-GCN~\cite{Xinshun} &  94.8 \\
    RE-TCN (Ours) & \textbf{96.34} \\
  \bottomrule
  \end{tabular}
\end{table}

\subsubsection{Robustness to Occlusion}
To evaluate the robustness of the proposed method against occlusion, we tested RE-TCN with occluded data from various perspectives. Our tests covered both spatial and temporal occlusions, reflecting scenarios likely to be encountered in real-world ambient-assisted living environments. For a fair comparison, we compared the performance of RE-TCN with state-of-the-art methods designed to handle occlusions. We designed our occlusion experiments following the same conditions described in~\cite{Song2020RichlyAG}, ~\cite{Zhenjie}, and~\cite{guo2024overcomplete} and compared our results with their methods. We used the NTU-RGB+D 60 X-sub dataset and defined three types of occlusions: frame, body part and random occlusion, as described in~\cite{Song2020RichlyAG},~\cite{Zhenjie} and~\cite{guo2024overcomplete}. The details of these occlusion types and the comparison results are discussed below. 

\noindent
\textbf{Frame Occlusion}

This occlusion type simulates temporal occlusion. As described in~\cite{Song2020RichlyAG} and~\cite{Zhenjie}, we randomly occluded consecutive sequences of frames from an action sequence. We set the length of the occluded consecutive frames to 10, 20, 30, 40, and 50. The experimental result is shown in Tab~\ref{tab:frame_occlusion_with_SOTA}. We conducted two experiments: one with a low probability and another with a high probability of frame occlusion occurring. The results reveal that methods not designed for occlusion robustness perform poorly as the occlusion length increases. Even methods designed to be robust against occlusion, such as~\cite{Song2020RichlyAG} and~\cite{Zhenjie}, showed a significant performance degradation with increased occlusion length. In contrast, our method demonstrates minimal performance degradation as occlusion length increases. Moreover, our results outperform existing methods across all the occlusion lengths, for both low and high probability scenarios. The performance gap between our method and those of~\cite{Song2020RichlyAG} and~\cite{Zhenjie} widens as the occlusion length increases, reaching about 49\% and 50.1\% respectively when the occlusion length is 50. 
\begin{table}[!htb]
    \caption{Experiment Results with Frame Occlusion on NTU-RGB+D 60 X-sub benchmark: The results are in accuracy (\%) with the best result highlighted in bold}
    \vspace{10pt}
    \label{tab:frame_occlusion_with_SOTA}
    \centering
    \footnotesize
    \setlength{\tabcolsep}{2.5pt}
    \begin{tabular}{lcccccc}
        \toprule
         Method & \multicolumn{6}{c}{Number of Occluded Frames} \\
         \cmidrule(lr){2-7}
          & 0 & 10 & 20 & 30 & 40 & 50 \\ \hline
         ST-GCN~\cite{yan2018spatial} & 80.7 & 69.3 & 57.0 & 44.5 & 34.5 & 24.0\\ 
         SR-TSL~\cite{Si2018SkeletonBasedAR} & 84.8 & 70.9 & 62.6 & 48.8 & 41.3 & 28.8\\ 
         STIGCN~\cite{Huang2020SpatioTemporalIG} & 88.8 & 70.4 & 51.0 & 38.7 & 23.8 & 8.0\\
         MS-G3D~\cite{Liu2020DisentanglingAU} & 87.3 & 77.6 & 65.7 & 54.3 & 41.9 & 30.1\\
         CTR-GCN~\cite{chen2021channel} & 87.5 & 72.4 & 54.1 & 35.6 & 22.4 & 11.5\\
         TCA-GCN~\cite{Wang2022SkeletonbasedAR} & \textbf{90.2} & 84.4 & 74.6 & 58.1 & 42.3 & 25.6\\
         HD-GCN~\cite{Yang2023HDGCNAH} & 86.8 & 57.0 & 29.5 & 18.5 & 11.2 & 7.04\\
         2s-AGCN~\cite{shi2019two} & 88.5 & 74.8 & 60.8 & 49.7 & 38.2 & 28.0\\ 
         1s RA-GCN~\cite{Song2020RichlyAG} & 85.8 & 81.6 & 72.9 & 61.6 & 47.9 & 34.0\\ 
         2s RA-GCN~\cite{Song2020RichlyAG}  & 86.7 & 83.0 & 76.4 & 65.6 & 53.1 & 39.5\\ 
         3s RA-GCN~\cite{Song2020RichlyAG}  & 87.3 & 83.9 & 76.4 & 66.3 & 53.2 & 38.5 \\ 
         1s PDGCN~\cite{Zhenjie}  & 85.7 & 81.9 & 75.4 & 66.4 & 54.9 & 40.0\\ 
         2s PDGCN~\cite{Zhenjie}  & 87.4 & 83.8 & 76.7 & 66.8 & 55.1 & 40.6\\ 
         3s PDGCN~\cite{Zhenjie}  & 87.5 & 83.9 & 76.6 & 66.7 & 53.9 & 40.0\\ 
         RE-TCN (P = 0.5) (Ours) & 89.33 & 87.68 & 86.75 & 84.18 & 85.13 & 81.24\\ 
         RE-TCN (P = 0.01) (Ours) & 89.85 & \textbf{89.77} & \textbf{89.75} & \textbf{89.68} & \textbf{89.67} & \textbf{89.60}\\ 
        \bottomrule
    \end{tabular}
\end{table}
\noindent

\textbf{Body Part Occlusion}

Body part occlusion aims to simulate scenarios where some body parts of a person are occluded by objects or self-occlusion, which is common in real-world unstructured environments. As defined in~\cite{Song2020RichlyAG} and~\cite{Zhenjie}, we occluded the left arm, right arm, two hands, two legs, and torso when testing the model. The experiment result is presented in Tab~\ref{tab:part_occlusion_with_SOTA}. The effects of occluding different body parts varied, with some parts showing better performance than others. Similar to frame occlusion, methods not designed to be robust against occlusion suffered performance degradation when body parts were occluded. For ~\cite{Song2020RichlyAG} and~\cite{Zhenjie}, the performance notably improved with body part occlusion. In comparison, our method outperforms existing methods across all body parts by a significant margin, validating the robustness of our method against occlusion.
\begin{table}[!htb]
    \caption{Experiment Results with Body Part Occlusion on NTU-RGB+D 60 X-sub benchmark: The results are in accuracy (\%) with the best result highlighted in bold}
    \vspace{10pt}
    \label{tab:part_occlusion_with_SOTA}
    \centering
    \footnotesize
    \setlength{\tabcolsep}{3pt}
    \begin{tabular}{lcccccc}
        \toprule
         Method & \multicolumn{6}{c}{Occlusion Body Parts} \\
         \cmidrule(lr){2-7}
          & None &  {\makecell{Left \\ Arm}} & {\makecell{Right \\ Arm}} & {\makecell{Two \\ Hands}} & {\makecell{Two \\ Legs}} & Trunk \\ \hline
         ST-GCN~\cite{yan2018spatial} & 80.7 & 71.4 & 60.5 & 62.6 & 77.4 & 50.2\\ 
         SR-TSL~\cite{Si2018SkeletonBasedAR} & 84.8 & 70.6 & 54.3 & 48.6 & 74.3 & 56.2 \\ 
         STIGCN~\cite{Huang2020SpatioTemporalIG} & 88.8 & 12.7 & 11.5 & 18.3 & 45.5 & 20.9\\
         MS-G3D~\cite{Liu2020DisentanglingAU} & 87.3 & 31.3 & 23.8 & 17.1 & 78.3 & 61.6\\
         CTR-GCN~\cite{chen2021channel} & 87.5 & 13.0 & 12.5 & 12.7 & 21.0 & 36.3\\
         TCA-GCN~\cite{Wang2022SkeletonbasedAR} & \textbf{90.2} & 75.4 & 53.4 & 70.8 & 75.2 & 78.6 \\
         HD-GCN~\cite{Yang2023HDGCNAH} & 86.7 & 67.1 & 55.7 & 56.7 & 74.8 & 61.3\\
         2s-AGCN~\cite{shi2019two} & 88.5 & 72.4 & 55.8 & 82.1 & 74.1 & 71.9\\ 
         1s RA-GCN~\cite{Song2020RichlyAG}  & 85.8 & 69.9 & 54.0 & 66.8 & 82.4 & 64.9\\ 
         2s RA-GCN~\cite{Song2020RichlyAG}  & 86.7 & 75.9 & 62.1 & 69.2 & 83.3 & 72.8\\ 
         3s RA-GCN~\cite{Song2020RichlyAG}  & 87.3 & 74.5 & 59.4 & 74.2 & 83.2 & 72.3\\ 
         1s PDGCN~\cite{Zhenjie}  & 85.7 & 73.4 & 60.4 & 65.9 & 83.0 & 71.2\\ 
         2s PDGCN~\cite{Zhenjie}  & 87.4 & 76.4 & 62.0 & 74.4 & 84.8 & 70.4\\ 
         3s PDGCN~\cite{Zhenjie}  & 87.5 & 76.0 & 62.0 & 75.4 & 85.0 & 74.3\\ 
         RE-TCN (Ours) & 89.78 & \textbf{89.11} & \textbf{88.60} & \textbf{88.26} & \textbf{89.46} & \textbf{89.29} \\ 
        \bottomrule
    \end{tabular}
\end{table}

\noindent
\textbf{Random Occlusion}

Random occlusion simulates how skeleton data might be obscured in real-world environments. Following the approach in~\cite{Song2020RichlyAG}, we set the occlusion probabilities to 0.2, 0.3, 0.4, 0.5, and 0.6 with the result shown in Tab~\ref{tab:Random_occlusion_with_SOTA}. We also compared our results with the approach in~\cite{guo2024overcomplete}, using their occlusion probabilities of 0.08, 0.1, 0.12, and 0.15, as presented in Tab~\ref{tab:random_oclusion_with_DAE}. 

Tab~\ref{tab:Random_occlusion_with_SOTA} reveals a significant performance drop as occlusion probabilities increase for those methods not designed to handle occlusion. While~\cite{Song2020RichlyAG} showed improved robustness, their method still suffers noticeable degradation at higher occlusion levels. Our method, however, experiences only a minimal performance loss as the probability increases and outperforms existing methods across all occlusion probabilities.

Tab~\ref{tab:random_oclusion_with_DAE} shows results for a denoising method used as a preprocessing module with state-of-the-art approaches. While this method maintains stable performance across various occlusion probabilities, our approach consistently outperforms it at all levels.

The superior performance of our method as demonstrated in both Tab~~\ref{tab:Random_occlusion_with_SOTA}, and~\ref{tab:random_oclusion_with_DAE}, underscores its robustness against occlusion. 
\begin{table}[!htb]
    \caption{Experiment Results with Random Occlusion on NTU-RGB+D 60 X-sub benchmark: The results are in accuracy (\%) with the best result highlighted in bold}
    \vspace{10pt}
    \label{tab:Random_occlusion_with_SOTA}
    \centering
    \footnotesize
    \begin{tabular}{lcccccc}
        \toprule
         Method & \multicolumn{6}{c}{Occlusion Probability} \\
         \cmidrule(lr){2-7}
          & 0 & 0.2 & 0.3 & 0.4 & 0.5 & 0.6 \\ \hline
         ST-GCN~\cite{yan2018spatial} & 80.7 & 12.4 & 6.6 & 6.2 & 4.0 & 4.2\\ 
         SR-TSL~\cite{Si2018SkeletonBasedAR} & 84.8 & 43.0 & 25.2 & 12.1 & 6.0 & 3.7\\   
         2s-AGCN~\cite{shi2019two} & 88.5 & 38.5 & 22.8 & 13.4 & 8.5 & 6.1\\ 
         RA-GCN~\cite{Liang} & 85.9 & 84.1 & 81.7 & 77.2 & 70.0 & 57.4\\
         1s RA-GCN~\cite{Song2020RichlyAG} & 80.0 & 75.1 & 68.4 & 57.4 & 44.7 & 27.6\\ 
         2s RA-GCN~\cite{Song2020RichlyAG}  & 82.5 & 79.7 & 76.2 & 71.0 & 62.0 & 48.7\\ 
         3s RA-GCN~\cite{Song2020RichlyAG}  & 82.7 & 79.8 & 75.6 & 68.9 & 58.1 & 43.7\\ 
         RE-TCN (Ours) & \textbf{88.66} & \textbf{88.63} & \textbf{88.43} & \textbf{88.40} & \textbf{88.60} & \textbf{88.21}\\ 
        \bottomrule
    \end{tabular}
\end{table}
\begin{table}[!htb]
    \caption{Experiment Results with Random Occlusion on NTU-RGB+D 60 X-sub benchmark: The symbol "w" denotes the use of DAE denoising method~\cite{guo2024overcomplete} and the results are in accuracy (\%) with the best result highlighted in bold}
    \vspace{10pt}
    \label{tab:random_oclusion_with_DAE}
    \centering
    \footnotesize
    \begin{tabular}{lccccc}
        \toprule
         Method & \multicolumn{5}{c}{Occlusion Probability} \\
         \cmidrule(lr){2-6}
          &  0 &  0.08 &  0.1 &  0.12 & 0.15\\ \hline
         EfficientGCN~\cite{song2022constructing}w\cite{guo2024overcomplete} & 87.74 & 87.66 & 87.66 & 87.57 & 87.51\\ 
         ST-GCN++~\cite{Haodong}w\cite{guo2024overcomplete} & 87.80 & 87.47 & 87.37 & 87.38 & 87.23\\ 
         CTR-GCN~\cite{chen2021channel}w\cite{guo2024overcomplete} & 89.20 & 89.19 & 89.19 & 89.12 & 88.12\\ 
         AAGCN~\cite{Hanqing}w\cite{guo2024overcomplete} & 88.71 & 88.42 & 88.42 & 88.37 & 88.42\\ 
         MS-G3D~\cite{Liu2020DisentanglingAU}w\cite{guo2024overcomplete} & 88.75 & 88.71 & 88.68 & 88.67 & 88.57\\ 
         RE-GCN (Ours) & \textbf{89.78}& \textbf{89.35} & \textbf{89.29} & \textbf{89.20} & \textbf{88.92}\\ 
        \bottomrule
    \end{tabular}
\end{table}

\subsubsection{Robustness to Noise}

This is designed to simulate the effect of noise in skeleton data, a common challenge in real-world environments. Following the approach described in~\cite{Song2020RichlyAG}, we designed two experiments with different Gaussian noises as shown in Tab~\ref{tab:Random_jittering_1_with_SOTA}, and~\ref{tab:Random_jittering_2_with_SOTA}. We set the probability for every joint to 0.02, 0.04, 0.06, 0.08, and 0.10. Additionally, we adopted the approach from~\cite{guo2024overcomplete} setting the jittering probability to 0.05, 0.1, 0.2 and 0.3 as shown in Tab~\ref{tab:random_jittering_with_DAE}.

In Tab~\ref{tab:Random_jittering_1_with_SOTA} and~\ref{tab:Random_jittering_2_with_SOTA}, we observe a significant performance degradation as the jittering probability increases, even with the method in~\cite{Song2020RichlyAG} designed to handle jittering in skeleton data. By contrast, our method demonstrates consistent performance across all the different probability levels, outperforming existing methods. Notably, for both tested values of $\sigma$, the performance gap between our method and the one in~\cite{Song2020RichlyAG} widens significantly as the jittering probability increases with a margin of 52.96\% and 27.57\% for $\sigma = 0.1$ and $\sigma = 0.05$ respectively at a probability of 0.1. 

In Tab~\ref{tab:random_jittering_with_DAE}, while the performance of the method in~\cite{guo2024overcomplete} demonstrates a stable performance across probability levels, our approach outperforms it across all the probability levels. 

Overall, the results in Tab~\ref{tab:Random_jittering_1_with_SOTA}, ~\ref{tab:Random_jittering_2_with_SOTA}, and~\ref{tab:random_jittering_with_DAE} highlight the robustness of our approach in mitigating the effects of jittering. 
\begin{table}[!htb]
    \caption{Experiment Results with Jittering Skeletons $(\mu = 0, \sigma = 0.1)$ on NTU-RGB+D 60 X-sub benchmark: The results are in accuracy (\%) with the best result highlighted in bold}
    \vspace{10pt}
    \label{tab:Random_jittering_1_with_SOTA}
    \centering
    \footnotesize
    \begin{tabular}{lcccccc}
        \toprule
         Method & \multicolumn{6}{c}{Jittering Probability} \\
         \cmidrule(lr){2-7}
          & 0 & 0.02 & 0.04 & 0.06 & 0.08 & 0.10 \\ \hline
         ST-GCN~\cite{yan2018spatial} & 80.7 & 66.4 & 44.1 & 32.7 & 13.3 & 7.0\\ 
         SR-TSL~\cite{Si2018SkeletonBasedAR} & 84.8 & 70.4 & 53.2 & 41.0 & 33.9 & 21.4\\   
         2s-AGCN~\cite{shi2019two} & 88.5 & 74.9 & 60.9 & 41.9 & 29.4 & 20.6\\ 
         RA-GCN~\cite{Liang} & 85.9 & 73.2 & 59.8 & 45.3 & 41.6 & 34.5\\
         1s RA-GCN~\cite{Song2020RichlyAG} & 85.8 & 84.1 & 66.1 & 34.2 & 22.2 & 13.9\\ 
         2s RA-GCN~\cite{Song2020RichlyAG}  & 86.7 & 70.0 & 55.3 & 48.2 & 41.5 & 36.4\\ 
         3s RA-GCN~\cite{Song2020RichlyAG}  & 87.3 & 84.2 & 72.4 & 61.6 & 42.4 & 28.7\\ 
         RE-TCN (Ours) & \textbf{89.58} & \textbf{89.47} & \textbf{89.49} & \textbf{89.45} & \textbf{89.22} & \textbf{89.36}\\ 
        \bottomrule
    \end{tabular}
\end{table}
\begin{table}[!htb]
    \caption{Experiment Results with Jittering Skeletons $(\mu = 0, \sigma = 0.05)$ on NTU-RGB+D 60 X-sub benchmark: The results are in accuracy (\%) with the best result highlighted in bold}
    \vspace{10pt}
    \label{tab:Random_jittering_2_with_SOTA}
    \centering
    \footnotesize
    \begin{tabular}{lcccccc}
        \toprule
         Method & \multicolumn{6}{c}{Jittering Probability} \\
         \cmidrule(lr){2-7}
          & 0 & 0.02 & 0.04 & 0.06 & 0.08 & 0.10 \\ \hline
         ST-GCN~\cite{yan2018spatial} & 80.7 & 76.4 & 65.1 & 50.2 & 32.8 & 19.5\\ 
         SR-TSL~\cite{Si2018SkeletonBasedAR} & 84.8 & 69.4 & 55.3 & 50.1 & 46.6 & 39.2\\   
         2s-AGCN~\cite{shi2019two} & 88.5 & 78.9 & 79.8 & 76.8 & 72.6 & 60.7\\ 
         RA-GCN~\cite{Liang} & 85.9 & 83.8 & 81.3 & 75.3 & 69.2 & 61.4\\
         1s RA-GCN~\cite{Song2020RichlyAG} & 85.8 & 82.4 & 77.1 & 72.3 & 63.8 & 49.9\\ 
         2s RA-GCN~\cite{Song2020RichlyAG}  & 86.7 & 83.8 & 77.3 & 71.6 & 61.6 & 58.5\\ 
         3s RA-GCN~\cite{Song2020RichlyAG}  & 87.3 & 87.0 & 84.5 & 81.1 & 72.9 & 61.4\\ 
         RE-TCN (Ours) & \textbf{89.12} & \textbf{89.04} & \textbf{89.01} & \textbf{88.99} & \textbf{88.90} & \textbf{88.97} \\ 
        \bottomrule
    \end{tabular}
\end{table}
\begin{table}[!htb]
    \caption{Experiment Results with Random Jittering on NTU-RGB+D 60 X-sub benchmark: The symbol "W" denotes the use of DAE denoising method~\cite{guo2024overcomplete} and the results are in accuracy (\%) with the best result highlighted in bold}
    \vspace{10pt}
    \label{tab:random_jittering_with_DAE}
    \centering
    \footnotesize
    \begin{tabular}{lccccc}
        \toprule
         Method & \multicolumn{5}{c}{Jittering Probability} \\
         \cmidrule(lr){2-6}
          & 0 &  0.05 &  0.1 &  0.2 & 0.3\\ \hline
         EfficientGCN~\cite{song2022constructing}w\cite{guo2024overcomplete} & 87.62 & 87.57 & 87.59 & 87.60 & 87.58\\ 
         ST-GCN++~\cite{Haodong}w\cite{guo2024overcomplete} & 87.89 & 87.80 & 87.75 & 87.58 & 87.31\\ 
         CTR-GCN~\cite{chen2021channel}w\cite{guo2024overcomplete} & 89.07 & 89.11 & 89.03 & 89.09 & \textbf{88.76}\\ 
         AAGCN~\cite{Hanqing}w\cite{guo2024overcomplete} & 88.57 & 88.59 & 88.65 & 88.65 & 88.45\\ 
         MS-G3D~\cite{Liu2020DisentanglingAU}w\cite{guo2024overcomplete} & 88.60 & 88.64 & 88.62 & 88.61 & 88.34\\ 
         RE-GCN (Ours) & \textbf{89.58} & \textbf{89.56} & \textbf{89.36} & \textbf{88.97} & 88.41\\ 
        \bottomrule
    \end{tabular}
\end{table}

\section{Per-Class Classification Performance}
\label{s:Per-class performance}
\noindent
We evaluated the classification performance for each class using NTU RGB+D 60, Northwestern-UCLA, SHREC'17, and DHG-14/28 datasets. For this evaluation, we computed the confusion matrix, as presented in Fig.~\ref{fig:confusion_matrix_figure} and classification report, which is provided in the supplementary material. As shown in the confusion matrix, the model achieved high accuracy across all classes, with the exception of the DHG-14/28 dataset, where the model occasionally misclassified the grab class as the pinch class. This confusion is reasonable given the high similarity between the two classes. Apart from this, the model consistently demonstrated high accuracy across all classes in the other datasets. This class-wise classification performance highlights the model's robustness and its ability to effectively handle diverse range of classes.

\begin{figure*}[h]
  \centering  \includegraphics[width=0.9\linewidth]{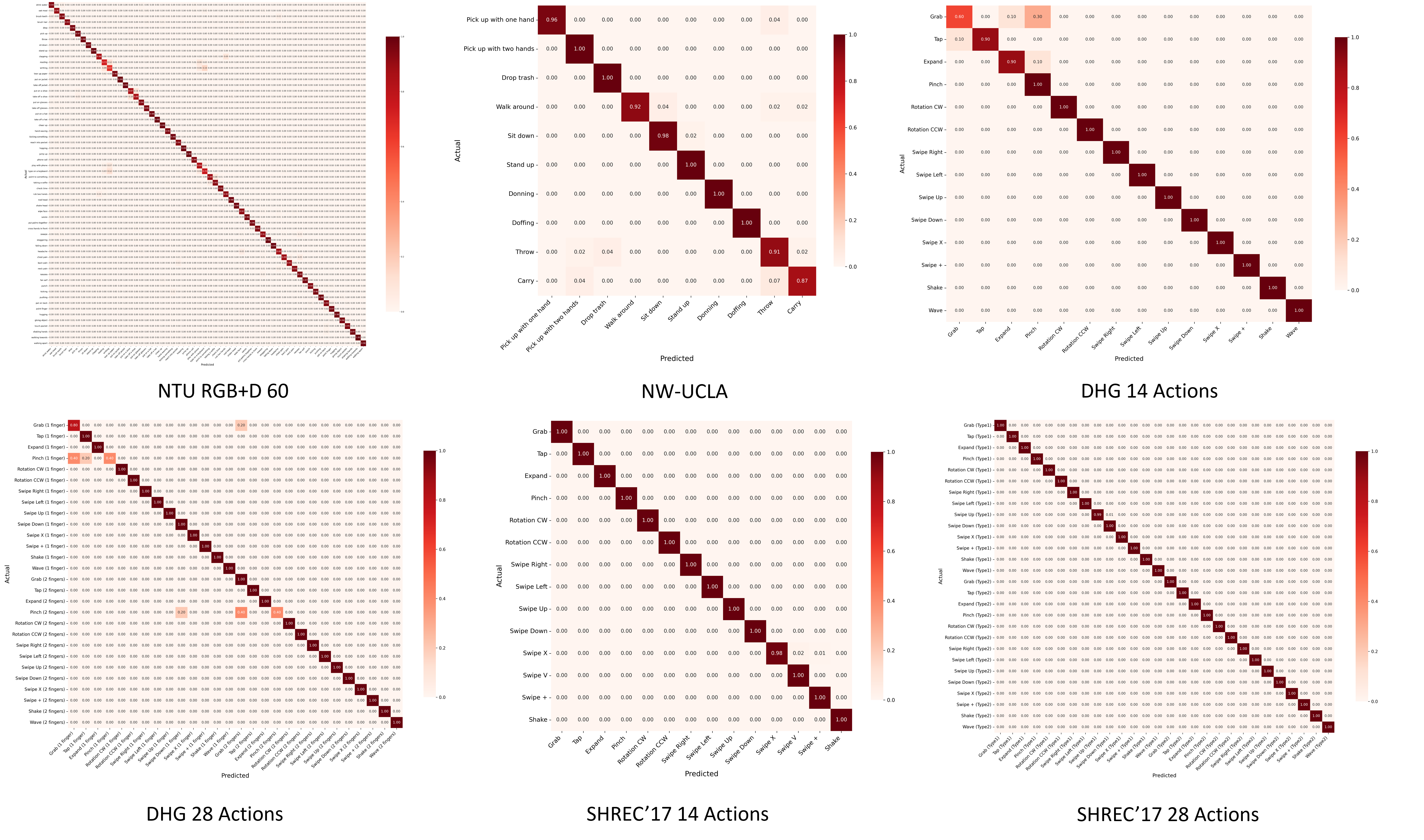}
  \caption{Confusion matrices showing classification performance for each class in the NTU RGB+D 60, Northwestern-UCLA, SHREC'17, and DHG-14/28 datasets} 
  \label{fig:confusion_matrix_figure}
\end{figure*}

\section{Real-time Application for Human Action Recognition}
\label{s:Application}
\begin{figure*}[h]
  \centering  \includegraphics[width=0.9\linewidth]{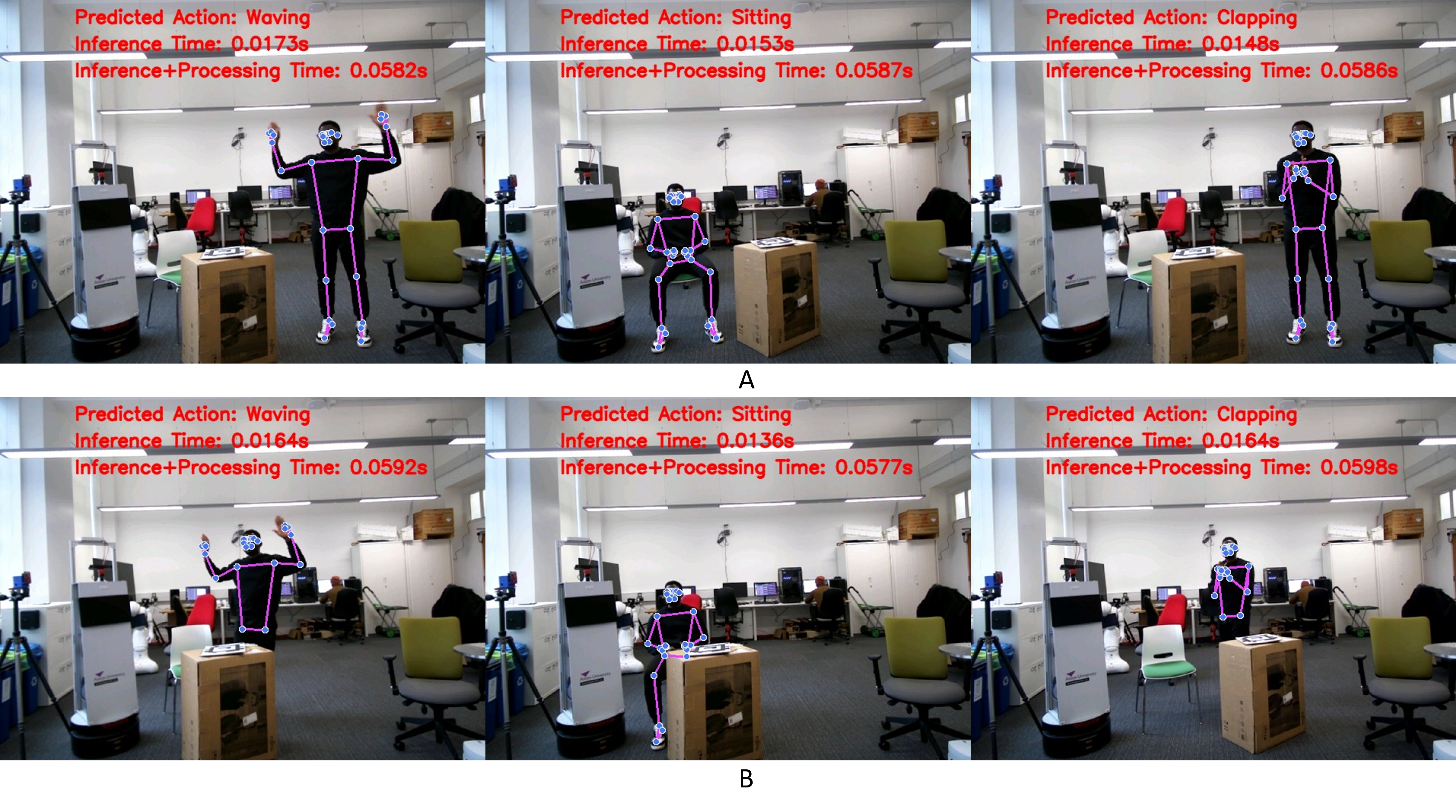}
  \caption{Real time action recognition: A) action recognition without occlusions B) action recognition with occlusions. \textbf{Predicted action}: The action recognised by the model. \textbf{Inference time}: The time taken for the model to generate a prediction. \textbf{Inference+Processing time}: The total processing time that spans from frame capture, pose extraction, and model prediction} 
  \label{fig:demo_figure}
\end{figure*}

\noindent
We developed a human action recognition system using RE-TCN to demonstrate its practical use in real-world, unstructured environments. The system was tested in a challenging environmental condition where objects partially block the camera's view, resulting in noisy data and occlusions. Using Mediapipe Pose~\cite{Bazarevsky2020BlazePoseOR}, the system processes RGB image sequences to estimate human poses, which then feeds into the model for real-time action recognition. We tested the system on a standard PC with an Intel Core i5 processor running Ubuntu 20.04 LTS, without a dedicated GPU. As shown in Fig.~\ref{fig:demo_figure}, the system successfully identifies actions in real time. The fast inference speed demonstrates that the model can perform effectively even on devices with limited resources. The model's reliable performance despite noise and occlusion demonstrates its suitability for practical applications. These results indicate strong potential for real-world deployment for ambient assisted living applications. 

\section{Conclusion and Future Studies}
\noindent
In this paper, we present RE-TCN, a model designed to address the challenges of noisy data, occlusion and computational cost in real-world ambient assisted living environments. The proposed RE-TCN model incorporated three key components: Adaptive Temporal Weighting, Depthwise Separable Convolutions, and data augmentation techniques. Through extensive experiments on the NTU-RGB+D 60, NW-UCLA, and SHREC'17 datasets, we demonstrate the effectiveness and robustness of the model. To evaluate the robustness of RE-TCN in the presence of noise and occlusions, we conducted tests simulating various real-world conditions. Across all test configurations, RE-TCN consistently achieves state-of-the-art performance. Additionally, the model exhibits superior accuracy and computational efficiency compared to existing approaches, indicating its potential for facilitating accurate action recognition in real-world ambient assisted living environments. 

Although the model achieved state-of-the-art performance on different testing configurations, there remain certain limitations that, if addressed, could further strengthen the model's applicability in practice. First, while RE-TCN demonstrates generalisability comparable to the current methods, there is a scope to optimise the architecture to improve its ability to handle data from individuals not represented during training. Secondly, although the datasets employed are comprehensive, using a dataset specifically collected from care homes or similar settings would more directly ensure that the model is applicable to its intended context. Thirdly, although the experiments accounted for various challenges likely to be encountered in real-world conditions, the model has yet to be deployed in a living setting. Field evaluations such as monitoring daily activities in care homes or the residences of elderly individuals could yield invaluable insights into its practical utility.  

Addressing these limitations will not only enhance RE-TCN adaptability and scalability but also further emphasise its relevance across a range of real-world environments.

\section{Acknowledgment}
The model was trained on the Aston EPS Machine Learning Server, funded by the EPSRC Core Equipment Fund, Grant EP/V036106/1.

% \bibliographystyle{IEEEtrans}
% \clearpage 
\bibliographystyle{unsrt}
\bibliography{main}
\vfill 
\mbox{} 
\end{document}